\pdfoutput=1

\documentclass[11pt]{article}

\usepackage[final]{acl}

\usepackage{times}
\usepackage{latexsym}

\usepackage[T1]{fontenc}

\usepackage[utf8]{inputenc}

\usepackage{microtype}

\usepackage{inconsolata}

\usepackage{graphicx}

\title{Learning Domain-Invariant Features for Out-of-Context News Detection}

\author{
 \textbf{Yimeng Gu\textsuperscript{1}},
 \textbf{Mengqi Zhang\textsuperscript{2}},
 \textbf{Ignacio Castro\textsuperscript{1}},
 \textbf{Shu Wu\textsuperscript{3}},
 \textbf{Gareth Tyson\textsuperscript{4}}
\\
\\
 \textsuperscript{1}Queen Mary University of London,
 \textsuperscript{2}Shandong University,
\\
 \textsuperscript{3}CRIPC, Chinese Academy of Science,
\\
 \textsuperscript{4}The Hong Kong University of Science and Technology (GZ)
\\
   \texttt{yimeng.gu@qmul.ac.uk}
}

\definecolor{babyblue}{rgb}{0.54, 0.81, 0.94}
\definecolor{mossgreen}{rgb}{0.68, 0.87, 0.68}
\definecolor{bananamania}{rgb}{0.98, 0.91, 0.71}
\usepackage{xspace}
\usepackage{todonotes}
\usepackage{multirow}
\usepackage{algorithm}
\usepackage{algpseudocode}
\usepackage{hyperref}
\usepackage{subfigure}
\usepackage{url}
\usepackage{graphicx}
\usepackage{amsmath}
\usepackage{booktabs}
\usepackage{adjustbox}
\usepackage{enumitem}
\usepackage{colortbl}
\usepackage{xcolor} 

\newcommand{\one}{({\em i}\/)\xspace}
\newcommand{\two}{({\em ii}\/)\xspace}
\newcommand{\three}{({\em iii}\/)\xspace}
\newcommand{\four}{({\em iv}\/)\xspace}

\def\ie{\emph{i.e.} \xspace}

\begin{document}
\maketitle
\begin{abstract}
Out-of-context news is a common type of misinformation on online media platforms. This involves posting a caption, alongside a mismatched news image. 
  Existing out-of-context news detection models only consider the scenario where pre-labeled data is available for each domain, failing to address the out-of-context news detection on unlabeled domains (e.g. news topics or agencies). 
  In this work, we therefore focus on \emph{domain adaptive} out-of-context news detection. 
  In order to effectively adapt the detection model to unlabeled news  \emph{topics} or \emph{agencies},
  we propose ConDA-TTA (Contrastive Domain Adaptation with Test-Time Adaptation) which applies contrastive learning and maximum mean discrepancy (MMD) to learn domain-invariant features. 
  In addition, we leverage test-time target domain statistics to further assist domain adaptation. 
  Experimental results show that our approach outperforms baselines in most domain adaptation settings on two public datasets, by as much as 2.93\% in F1 and 2.08\% in accuracy. 
\end{abstract}

\section{Introduction}
\label{sec:introduction}
\begin{figure}[htbp]
  \centering
  \includegraphics[width=0.5\textwidth]{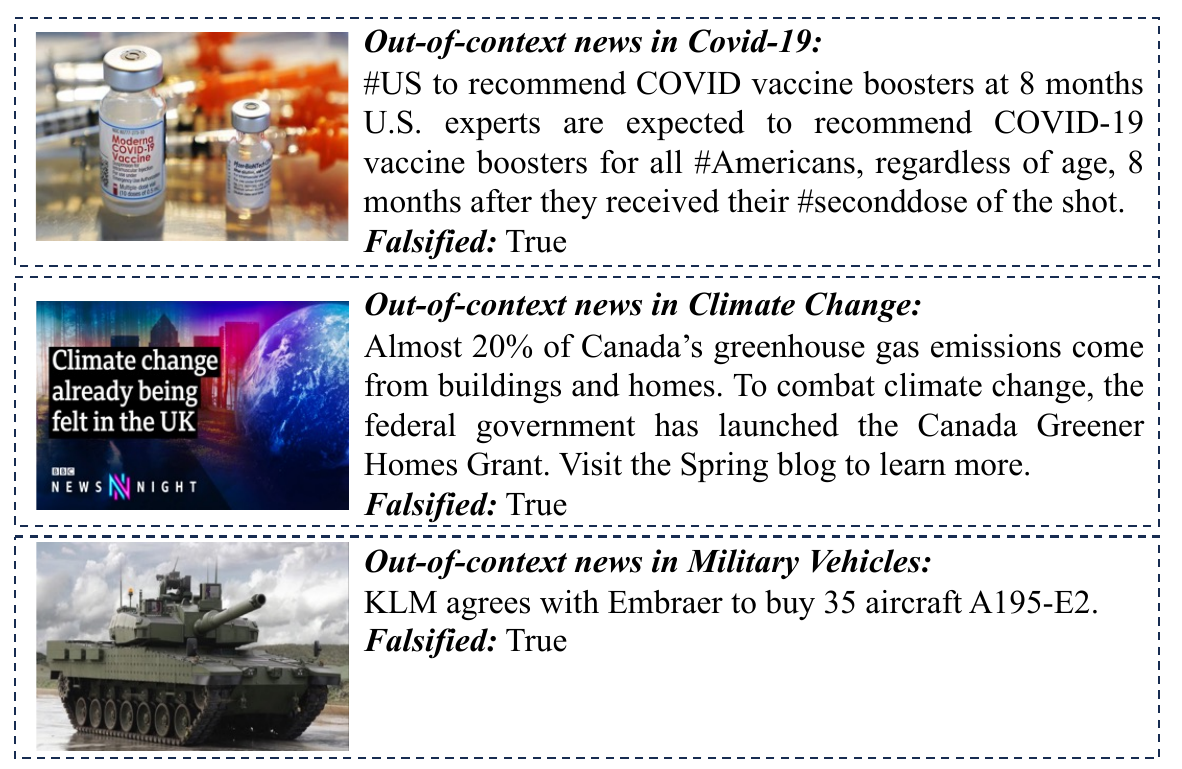}
  \caption{Examples of out-of-context news of three different news topics from the Twitter-COMMs dataset.}
  \label{fig:ooc-example}
  \vspace{-0.2cm}
\end{figure}

Online news platforms suffer from the release and spread of misinformation.
A common~\cite{fazio2020ooc} yet subtle type of misinformation is image repurposing, also called \emph{out-of-context news}. 
Unlike DeepFake~\citep{dolhansky2020deepfake} models which generate non-existing images, this approach attaches a false claim with a real image outside of its original context (see Figure~\ref{fig:ooc-example}).
This lowers the technical threshold~\cite{fazio2020ooc} of producing misinformation since anyone can trivially attach a fabricated caption to an image and post it online, thereby enlarging its scope of occurrence.

Out-of-context news detection has received a growing
attention in recent years. Existing works~\citep{biamby-etal-2022-twitter, luo-etal-2021-newsclippings, abdelnabi2022open, mu2023self} fine-tune pre-trained vision and text encoders to represent the image and text.
More recent works~\cite{dai2024instructblip, zhang2023detecting} query Multimodal Large Language Models (MLLMs) to obtain the prediction with explanations. 
However, 
their approaches do not address the model adaptation to entirely new topics or news agencies. 
Due to the wide array of news topics shared online, it is infeasible to re-train the detector every time a new topic emerges. 
Considering annotation costs, it is equally critical to build a detection model that can adapt to other news agencies with minimal efforts.

Unfortunately, prior works neglect addressing the domain adaption issue on out-of-context news detection. 
They tend to capture domain-specific knowledge not shared across different domains.
Although this knowledge is helpful for improving the model's performance on annotated domains, it results in inferior performance on unannotated domains~\cite{wang2018eann}. In light of this, we believe that learning domain-invariant features will help the detection model adapt to unlabeled domains.
However, this comes with several challenges.
\emph{First}, 
existing out-of-context news datasets only cover a small number of topics/agencies. This makes it challenging for the detection model to learn domain-invariant features, because the learned features might be biased towards the limited number of source domains. 
\emph{Second}, in (social media) news dataset, the data distributions of different domains vary, since news from different domains tend to have different writing styles and image styles. This makes domain-invariant knowledge learning challenging. It is critical to ensure that transferable domain-invariant knowledge is actually learned. Otherwise, the model may bypass learning it and end up learning semantic patterns as a shortcut to fulfill the classification task~\citep{li2023rethinking, chi2023adapting}. 

To tackle the aforementioned challenges, we propose \emph{Contrastive Domain Adaptation with Test-Time Adaptation} (ConDA-TTA) for domain adaptive out-of-context news detection.
ConDA-TTA first uses an MLLM to directly encode the image and text into a multimodal feature representation.
After that, we adopt the contrastive loss to learn a representation in the projected space where the original news stays further away from the out-of-context news. To overcome the shortage of labeled source domains, we take advantage of the unlabeled target domain data as well, to learn less biased domain-invariant features.
To ensure that domain-invariant features are captured from the learned representations, we then apply Maximum Mean Discrepancy (MMD) to reduce the discrepancy of the learned representations between the source and target domain. 
Finally, we incorporate Test-Time Adaptation (TTA) to update the statistic-related model parameters in the evaluation phase so that the model can better adapt to the target domain.

To summarize, our main contributions are as follows:

\begin{itemize}[noitemsep,nolistsep]
\item
We are the first to investigate domain adaptation in out-of-context news detection. To address the challenges, we propose a novel approach that learns domain-invariant features through MMD and TTA. 
\item
We evaluate our approach and demonstrate its effectiveness. 
Notably, it outperforms the baseline by as much as 2.93\% in F1 when the domain is defined as news topic. 
Additionally, it outperforms the baseline by as much as 1.82\% in F1 when the domain is defined as news agency.
\item
We conduct a comprehensive ablation study and show that MMD is the most contributing component to the domain adaptation on Twitter-COMMs, while TTA is the most contributing component on NewsCLIPpings. 
\end{itemize}

\section{Related Work}

\label{sec:related-work}

\subsection{Out-of-Context News Detection}
\label{sec:ooc}

Prior works on out-of-context news detection mainly adopt three technical routes: \one fine-tuning or \two prompting large vision and language models, and \three leveraging synthetic information.
\citet{biamby-etal-2022-twitter} fine-tunes CLIP~\citep{radford2021learning} to detect out-of-context news. Similarly, \citet{luo-etal-2021-newsclippings} fine-tunes both CLIP and VisualBert~\citep{li2019visualbert}. \citet{abdelnabi2022open} additionally leverages the retrieved textual and visual evidences. These works involve fine-tuning large multimodal pre-trained models, which is very compute-intensive and does not enable domain adaptation. 

Later work leverages synthetic multimodal information. \citet{shalabi2023image} generates augmented image from text, and generates augmented text from image. \cite{yuan2023support} extracts stances from external multimodal evidences to enhance the detection. 
More recent works focus on enhancing interpretability. \cite{qi2024sniffer} adopts two-stage instruction tuning on Instruct-BLIP~\cite{dai2024instructblip} to provide accurate and persuasive explanations to the prediction. \cite{zhang2023detecting} uses neural symbolic model to enhance model interpretability. \citet{shalabi2024leveraging} fine-tunes MiniGPT-4 to detect out-of-context news. Although effective, the above works do not address the domain adaptation problem in out-of-context news detection.

\subsection{Domain Adaptive Fake News Detection}
\label{sec:da}
To the best of our knowledge, no prior work has explored domain adaptive out-of-context news detection. However, existing works have investigated domain adaptation in fake news detection --- a task close to out-of-context news detection. Many works~\cite{wang2018eann, 9206973, 9285217, YUAN2021113633} adopt adversarial learning to learn the domain-invariant feature. 
Furthermore, by incorporating user comments and user-news interaction information, \citet{mosallanezhad2022domain} adopts reinforcement learning that exploits cross-domain and within-domain knowledge to achieve robustness in the target domain. \citet{yue2022contrastive} proposes a contrastive domain adaptation method in order to reduce the intra-class discrepancy and enlarge the inter-class discrepancy. \citet{lin2023zero} employs prompt engineering to learn language-agnostic contextual representations and models the domain-invariant structural features from the propagation threads. 

In contrast, our work focuses on the domain adaptation in out-of-context news detection and adopts both contrastive learning and test-time adaptation to address the unique challenges (Section ~\ref{sec:introduction}) of this task. 

\section{Problem Statement}
\label{sec:problem-statement}

In out-of-context news detection, given the news image-text pair $(x_{img}, x_{txt})$, the detection model is expected to predict whether it is falsified, \ie out-of-context. If the image and text are not from the same news post, this news is considered as falsified (labeled as \texttt{True}), otherwise not falsified (labeled as \texttt{False}).

In this work, we tackle the challenges of domain adaptive out-of-context news detection. During training, we have access to the labeled source domain data and unlabeled target domain data. During testing, we evaluate our model's performance on the labeled target domain data. 
The source domain and target domain are mutually exclusive. 
Concretely, the dataset is comprised of news from $N$ domains. 
Each time we select $M$ domains as the target domain, $\mathcal{D}^{T}=\{\mathcal{D}^{T_{m}}\}_{m=1}^{M}$, and the rest $N-M$ domains as the source domain, $\mathcal{D}^{S}=\{\mathcal{D}^{S_{n}}\}_{n=1}^{N-M}$.
The $n$-th source domain $\mathcal{D}^{S_{n}}$ contains labeled data, $\mathcal{D}^{S_{n}}=\{(x_{i}^{S_{n}}, y_{i}^{S_{n}})\}_{i=1}^{|S_{n}|}$, where $x^{S_{n}}_{i} = (x_{img}, x_{txt})_{i}^{S_{n}}$ denotes the $i$-th news image-text pair within domain $\mathcal{D}^{S_{n}}$, and $y_{i}^{S_{n}}$ denotes whether this image-text pair is falsified or not. In contrast, the $m$-th target domain contains unlabeled data: $\mathcal{D}^{T_{m}}=\{x_{j}^{T_{m}}\}_{j=1}^{|T_{m}|}$, where $x_{j}^{T_{m}} = (x_{img}, x_{txt})_{j}^{T_{m}}$. 

For simplicity, in the following notations, we use $T$ to denote target domain and $S$ to denote source domain.

\section{Our Approach: ConDA-TTA}
\label{sec:our-approach}

\begin{figure*}[htbp]
  \centering
  \includegraphics[width=1.0\textwidth]{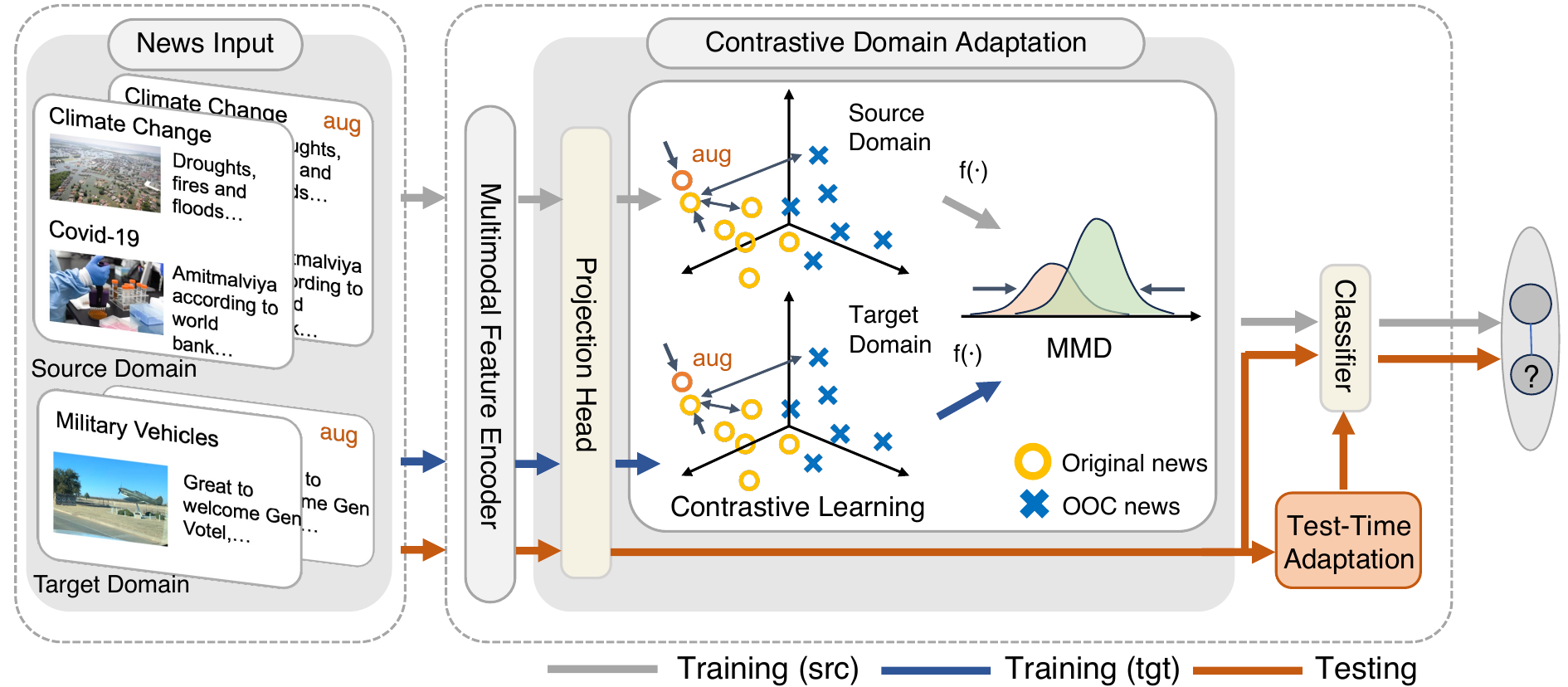}
  \caption{The model architecture of ConDA-TTA. We first use the \one \emph{Multimodal Feature Encoder} to encode the news and its augmentation into multimodal representations. Then in the \two \emph{Contrastive Domain Adpatation}, we apply contrastive learning and maximum mean discrepancy (MMD) to learn the domain-invariant feature. Finally, we adopt the \three \emph{Test-Time Adaptation} to update statistic-related model parameters in the evaluation phase.}
  \label{fig:arch}
  \vspace{-0.2cm}
\end{figure*}

In this section, we describe our proposed approach for domain adaptive out-of-context news detection: \emph{Contrastive Domain Adaptation with Test-Time Adaptation} (ConDA-TTA). The overall architecture of ConDA-TTA is illustrated in Figure~\ref{fig:arch}.
It consists of three components: \one \emph{Multimodal Feature Encoder}, which encodes the image and text pair into a multimodal representation; \two \emph{Contrastive Domain Adaptation}, which applies contrastive learning to learn a more separable representation space and maximum mean discrepancy (MMD) to learn the domain-invariant feature; and \three \emph{Test-Time Adaptation}, which leverages unlabeled test set statistics to further adapt the model to the target domain. 

\subsection{Multimodal Feature Encoder}
\label{sec:multimodal-feature-encoder}

In order to capture the semantic differences between original news and out-of-context news, we use MLLM to directly encode the news text and image into a meaningful multimodal representation. 


Formally, the multimodal feature encoder is defined as follows:
\begin{equation}
\begin{aligned}
\textbf{x}^{S} & = \mathrm{MLLM}((x_{img}, x_{txt})^{S}),\\
\textbf{x}^{T} & = \mathrm{MLLM}((x_{img}, x_{txt})^{T}),
\end{aligned}
\end{equation}
where $\textbf{x}^{S}$ and $\textbf{x}^{T}$ respectively denote the multimodal feature representation of the news image-text pair $(x_{img}, x_{txt})$ from the source and target domain.

\subsection{Contrastive Domain Adaptation}
\label{sec:domain-invariant}

After encoding the news, we introduce the contrastive domain adaptation module.
Specifically, this module first adopts contrastive learning to learn a more separable representation space for news image-text pairs and then uses MMD to capture the invariant features among different domains.

We first use a projection head to project the multimodal feature representation into a lower dimensional space. 
The projection head can be expressed as follows: 
\begin{equation}
\label{eq:proj}
\textbf{z}^{S} = \mathrm{Projection}(\textbf{x}^{S}), \quad \textbf{z}^{T} = \mathrm{Projection}(\textbf{x}^{T}).
\end{equation}

Next, we aim to train the projection head so that it can learn a more separable representation space and capture the domain-invariant features from $\textbf{x}^{S}$ and $\textbf{x}^{T}$ afterwards. 
This is achieved by applying the contrastive learning and maximum mean discrepancy respectively.
Note, no label information is required within this module.

\textbf{Contrastive Learning:} Contrastive learning has been shown effective in learning better representations for the classification
task~\cite{qian2022contrastive}. Inspired by~\cite{bhattacharjee-etal-2023-conda}, we leverage contrastive learning to learn more separable representations of the input multimodal features in the projected space. In this way, we expect that the learned representations of similar semantic meanings stay closer, and the learned representations of dissimilar semantic meanings stay further apart, so that it could make the classification easier.

To achieve this, we adopt the contrastive loss used in ~\cite{bhattacharjee-etal-2023-conda}. Applying contrastive loss pulls closer the \emph{positive pairs} and pushes further apart the \emph{negative pairs}. To construct the positive and negative pairs, we first augment each news item. For each news item $\textbf{x}_{i}^{S}$ in a source domain training batch $\{\textbf{x}_{i}^{S}\}_{i=1}^{b}$, we generate one augmentation of it and encode it by the same multimodal feature encoder to obtain its representation, denoted as $\textbf{x}_{i+}^{S}$. In this way, we obtain $2|b|$ news items in the augmented batch. After projecting the input multimodal feature to the lower dimensional space, the \emph{positive pair} is formed by the data item and its corresponding augmentation item, denoted as $(\textbf{z}_{i}^{S}, \textbf{z}_{i+}^{S})$; the \emph{negative pairs} are formed by the data item and the rest of the $2(|b|-1)$ items within the augmented batch, denoted as $(\textbf{z}_{i}^{S}, \textbf{z}_{k}^{S})$. The contrastive loss is expressed as follows:
\begin{equation}
\label{eq:contrastive-loss}
\mathcal{L}_{ctr}^{S} = - \sum_{(i, i_{+}) \in b} \log \frac{\exp(\mathrm{sim}(\textbf{z}_{i}^{S}, \textbf{z}_{i_{+}}^{S})/t)}{\sum_{k=1, k \ne i}^{2|b|} \exp(\mathrm{sim}(\textbf{z}_{i}^{S}, \textbf{z}_{k}^{S})/t)}.
\end{equation}

Here, $b$ denotes the current batch and $|b|$ denotes the size of the current batch. $S$ denotes the source domain. $\textbf{z}_{i}^{S}$ and $\textbf{z}_{j}^{S}$ denote the learned representations of the $i$-th data item and its augmentation in the projected space. $t$ denotes the temperature coefficient. $\mathrm{sim}(\cdot)$ denotes the similarity metric. 
Here, cosine similarity is being used. The data augmentation on the target domain training batch and the contrastive loss computation for the target domain $\mathcal{L}_{ctr}^{T}$ are applied in the same way. Note, the augmentation preserves the label. 


\textbf{Maximum Mean Discrepancy (MMD):} Now that we have learned more separable representations, $\textbf{z}^{S}$ and $\textbf{z}^{T}$, in the projected space, we want to further capture domain-invariant features. 
To achieve this, we adopt maximum mean discrepancy (MMD)~\cite{gretton2012kernel}. MMD is a kernel-based statistical test used to determine if two samples are drawn from different distributions~\cite{gretton2012kernel}.
We use it as a loss function to measure the discrepancy between two distributions.
Here, we regard the learned representations $\textbf{z}^{S}$ and $\textbf{z}^{T}$ as the samples from two random variables $Z^{S}$ and $Z^{T}$, and we want their probability distributions to have as small as possible discrepancy. In this way, the distribution of $Z^{S}$ and $Z^{T}$ conditions less on a specific domain. Alternatively speaking, if MMD is small enough, we regard that it removes the domain-specific features from $\textbf{z}$ and keeps the domain-invariant features. 

The squared MMD between $Z^{S}$ and $Z^{T}$ in the reproducing kernel Hilbert space (RKHS) $\mathcal{H}$ is formulated as follows:
\begin{equation}
\begin{aligned}
\label{eq:mmd}
\mathrm{MMD}^{2} & = \left[ \sup_{||f||_{\mathcal{H}}\le1} (\mathbf{E}[f(Z^{S})] - \mathbf{E}[f(Z^{T})])\right]^{2} \\
& = ||\mu_{S} - \mu_{T}||^2_{\mathcal{H}},
\end{aligned}
\end{equation}
where $\mu_{S}$ and $\mu_{T}$ are mean embeddings of $f(Z^{S})$ and $f(Z^{T})$, $f\in\mathcal{F}$. $\mathcal{F}$ is the unit ball function class in the Hilbert space $\mathcal{H}$.

Following prior work~\citep{pan2010domain, long2015learning, bhattacharjee-etal-2023-conda}, we compute the MMD of features in the lower dimensional space (\ie the projected $\textbf{z}$ space). In our implementation, we use the empirical MMD~\cite{gretton2012kernel}, which is expressed as follows:
\begin{equation}
\begin{aligned}
\label{eq:empirical-mmd}
\mathrm{MMD}  = &\left[\frac{1}{m^{2}}\sum_{i, j=1}^{m}k(\textbf{z}_{i}^{S}, \textbf{z}_{j}^{S}) - \frac{2}{mn}\sum_{i, j=1}^{m,n}k(\textbf{z}_{i}^{S}, \textbf{z}_{j}^{T}) \right.\\
& \left. + \frac{1}{n^{2}}\sum_{i, j=1}^{n}k(\textbf{z}_{i}^{T}, \textbf{z}_{j}^{T})\right]^{\frac{1}{2}}.
\end{aligned}
\end{equation}

Here, both $m$ and $n$ are equal to the batch size. $k(\cdot, \cdot)$ denotes the kernel function. $\sigma$ is a free parameter. We use the Radial basis function kernel shown in the following:
\begin{equation}
\label{eq:rbf}
k(\textbf{z}_{i}, \textbf{z}_{j}) = \exp \left( - \frac{||\textbf{z}_{i} - \textbf{z}_{j}||^{2}_{2}}{2\sigma^{2}}\right).
\end{equation}

\textbf{Training:} The learned domain-invariant representation, $\textbf{z}$, is then input to the classifier to get the predicted label $\hat{y}$ for the corresponding out-of-context news:
\begin{equation}
\hat{y} = \mathrm{CLS}(\textbf{z}).
\end{equation}

The loss function for the classifier is the cross-entropy loss, where $\hat{y}$ is the predicted label and $y$ is the ground true label:
\begin{equation}
\mathcal{L}_{CE} = - \mathbf{E}_{y \sim Y}\left[y\log(\hat{y}) + (1-y)\log(1-\hat{y})\right].
\end{equation}

During training, both the projection head and the classifier have learnable parameters. The total loss is composed of \one the contrastive losses from both the source and target data in the training set, \two the MMD between the source and target projected representations, and \three the cross-entropy losses of the predicted label for both the source data ($\mathcal{L}_{CE}$) and its augmentation ($\mathcal{L}_{CE+}$). This can be expressed as follows:
\begin{equation}
\label{eq:total-loss}
\begin{aligned}
\mathcal{L} = & \frac{1}{2}\lambda_{CE} (\mathcal{L}_{CE}^{S} + \mathcal{L}_{CE+}^{S}) + \frac{1}{2} \lambda_{ctr} (\mathcal{L}_{ctr}^{S} + \mathcal{L}_{ctr}^{T}) \\
& + \lambda_{MMD}\mathrm{MMD},
\end{aligned}
\end{equation}
where $\lambda_{CE}$ is the weight of the cross entropy losses, $\lambda_{ctr}$ is the weight of the contrastive losses and $\lambda_{MMD}$ is the weight of MMD.

\subsection{Test-Time Adaptation}

Domain-invariant features help the model generalize to the target domain. In this section, we introduce test-time adaptation which leverages target domain statistics to further adapt the model to the target domain during evaluation.


Prior work~\cite{li2016revisiting} observes that the statistics of the Batch Normalization (BN) layer contain domain-specific information. Concretely, they find that running test set through the trained model achieves deep adaptation effects on the target domain. The reason is that this process allows BN statistics to keep track of the target domain statistics, which is conducive to the domain adaptation.

At each training step, BN statistics compute and update the running estimates $\{\hat{\mu}_{k+1}, \hat{\sigma}^{2}_{k+1}\}$ (Eq.~\eqref{eq:stat-estimate}) based on its observed mean $\mu_{k}$ and variance $\sigma_{k}^{2}$ (Eq.~\eqref{eq:stat-compute}) of the layer batch inputs $\{x_{i}\}_{i=1}^{b}$:
\begin{equation}
\label{eq:stat-estimate}
\hat{\mu}_{k+1} = (1 - \rho) \hat{\mu}_{k} + \rho \mu_{k}, \quad  \hat{\sigma}^{2}_{k+1} = (1 - \rho) \hat{\sigma}^{2}_{k} + \rho \sigma^{2}_{k},
\end{equation}
\begin{equation}
\label{eq:stat-compute}
\mu_{k} = \frac{1}{b}\sum_{i}x_{i}, \quad
\sigma^{2}_{k} = \frac{1}{b}\sum_{i}(x_{i}-\mu_{k})^{2}.
\end{equation}
Here, $k$ denotes the $k$-th training step, and $\rho$ is the momentum with a default value of 0.1. $x_{i}$ denotes the input to BN. BN statistics are then fixed (denoted as $\mu$ and $\sigma^{2}$) and used for normalization when the model is set to evaluation mode~\cite{pytorch-bn}. 

As such, we add BN layers into the classifier. 
To integrate target domain statistics into BN's memory, we first pass the unlabeled target domain test data through the classifier before setting it to the evaluation mode. 
After that, we set the classifier to evaluation mode and evaluate it on the test set. 
In this way, the running estimates would keep the target domain statistics as well into record, which is helpful for the domain adaptation. 

\section{Experimental Design}
\label{sec:exp}

\subsection{Datasets}
\label{sec:datasets}

We conduct experiments on Twitter-COMMs and NewsCLIPpings, which are the only two out-of-context news datasets that conform to our problem definition (Section \ref{sec:problem-statement}).


\textbf{Twitter-COMMs~\cite{biamby-etal-2022-twitter}:} This dataset is collected from Twitter solely on three topics: \textit{Covid-19} (Cv), \textit{Climate Change} (Cl) and \textit{Military Vehicles} (M). 
After downloading the tweets and removing the tweets that has an unavailable image, we obtain 2,143,934 items (Cv: 1,387,043, Cl: 512,490, M: 244,401) in the training set, and 22,082 items (Cv: 6,456, Cl: 8,488, M: 7,138) in the test set. The labels are balanced. 

\textbf{NewsCLIPpings~\cite{luo-etal-2021-newsclippings}:} This dataset is derived from the VisualNews dataset~\citep{liu2020visual}, a benchmark news image captioning dataset. The news is collected from four agencies, namely \textit{Guardian} (G), \textit{BBC} (B), \textit{Washington Post} (W) and \textit{USA Today} (U). 
After downloading the dataset, we obtain a total number of 1,182,900 items (B: 205,968, G: 567,012, U: 228,393, W: 181,527) in the training set, and 124,168 items (B: 22,074, G: 59,163, U: 24,243, W: 18,688) in the test set. The labels are balanced.

\subsection{Implementation Details}
\label{sec:implementation-details}

\textbf{Augmentation:}
In our approach, we apply Gaussian blur to the image while keeping the text unaltered to generate the augmentation of the anchor item. More details and experimental results can be found in Appendix~\ref{sec:appendix-implementation}.

\textbf{Architecture:} 
We use BLIP-2's multimodal feature extractor to embed the image and text, and obtain the embedding $\textbf{x}$ of size 768. Details of multimodal feature encoder selection can be found in Appendix~\ref{sec:appendix-implementation}.
The projection head is composed of two linear layers with 768 and 500 (experimented with \{300, 500, 768\}) neurons respectively. The classifier is composed of three linear layers with 768, 768 and 2 neurons respectively, and two batch normalization layers following the first two linear layers. We use \texttt{Tanh()} as the activation function. We apply dropout both before the first and the third linear layer. The dropout rate is set to 0.2 (we experiment with \{0.2, 0.5\} and without dropout).

\textbf{Training:} We perform training using the Adam optimizer with a batch size of 256 for 20 epochs. The early stopping epoch is set to 5. The learning rate is set to 2e-4 for Twitter-COMMs and 1e-4 for NewsCLIPpings. For the loss function, $\lambda_{MMD}$ is set to 1 and both $\lambda_{CE}$ and $\lambda_{ctr}$ are set to 0.5.


\textbf{Source and Target Domain Partition:} 
We treat news toics as individual domains in Twitter-COMMs.
We use one topic as the target domain and the remaining two topics as the source domain. 
We treat news agencies as individual domains in NewsCLIPpings.
Based on an observation~\citep{liu2020visual} on the correlation of these four news agencies and our pilot experiments, we use two news agencies of the same country as the target domain and the rest two as the source domain respectively. 

\subsection{Baseline Models}
\label{sec:baseline-models}

\begin{table*}
  \caption{Evaluation of domain adaptation performance on TwitterCOMMs and NewsCLIPpings. Cv, Cl and M stand for covid-19, climate change and military vehicles respectively. B, G, U and W stand for BBC, Guardian, USA Today and Washington Post respectively. In X $\rightarrow$ Y, X denotes the source domain, Y denotes the target domain. The best performance is in \textbf{bold text}. The second best performance is \underline{underlined}.}
  \label{tab:eval}
  \resizebox{\textwidth}{!}{\begin{tabular}{lp{0.5cm}p{0.5cm}p{0.5cm}p{0.5cm}p{0.5cm}p{0.5cm}p{0.5cm}p{0.5cm}p{0.5cm}p{0.5cm}p{0.5cm}p{0.5cm}p{0.5cm}p{0.5cm}}
    \toprule
     \multirow{3}{*}{Model} & \multicolumn{6}{c}{Twitter-COMMs} & \multicolumn{8}{c}{NewsCLIPpings}\\
     ~ & \multicolumn{2}{c}{Cv, Cl $\rightarrow$ M} & \multicolumn{2}{c}{Cv, M $\rightarrow$ Cl} & \multicolumn{2}{c}{Cl, M $\rightarrow$ Cv} & \multicolumn{2}{c}{U, W $\rightarrow$ B} & \multicolumn{2}{c}{U, W $\rightarrow$ G} & \multicolumn{2}{c}{B, G $\rightarrow$ U} & \multicolumn{2}{c}{B, G $\rightarrow$ W} \\
    \cmidrule(lr){2-3} \cmidrule(lr){4-5} \cmidrule(lr){6-7} \cmidrule(lr){8-9} \cmidrule(lr){10-11} \cmidrule(lr){12-13} \cmidrule(lr){14-15}
    ~ & F1 & Acc & F1 & Acc & F1 & Acc & F1 & Acc & F1 & Acc & F1 & Acc & F1 & Acc \\
    \midrule
    Source only & 70.00 & 72.36  & 78.48 & 78.83  & 76.92 & 77.87 & 67.19 & 67.90 & 75.08 & 75.23 & 79.77 & 79.94 & 78.60 & 78.67 \\
    Source and target & 80.60 & 80.68 & 81.47 & 81.48 & 81.50 & 81.58 & 73.80 & 74.71 & 76.00 & 77.96 & 82.70 & 84.37 & 79.80 & 80.74 \\
    \cmidrule{1-15}
    EANN & 76.33 & \underline{77.26} & 79.13 & 79.16 & 79.44 & 79.60 & 69.30 & 69.61 & \textbf{76.01} & \textbf{76.12} & 79.72 & 79.88 & 78.27 & 78.34 \\
    MDA-WS & \underline{76.83} & 77.09 & 76.12 & 76.99 & 78.39 & 78.55 & 69.62 & 69.79 & 74.88 & 74.89 & 79.54 & 79.73 & 78.22 & 78.23\\
    CANMD & 67.13 & 68.16 & \underline{79.87} & \underline{80.53} & \underline{79.91} & \underline{80.01} & 68.42 & 68.62 & 74.75 & 75.64 & 79.30 & 79.69 & \textbf{79.93} & \textbf{78.59}\\
    REAL-FND & 73.50 & 74.82 & 79.13 & 79.36 & 76.64 & 77.62 & \underline{69.70} & \underline{69.96} & 75.20 & 75.30 & 79.95 & 80.00 & \underline{78.50} & \underline{78.52}\\
    CADA & 76.17 & 74.92 & 79.12 & 79.11 & 79.55 & 79.62 & 69.67 & 69.36 & 75.67 & \underline{75.86} & \underline{80.20} & \underline{80.07} & 78.28 & 78.23\\
    \midrule
    ConDA-TTA & \textbf{79.26} & \textbf{79.34} & \textbf{80.80} & \textbf{80.80} & \textbf{80.06} & \textbf{80.08} & \textbf{71.52} & \textbf{71.80} & \underline{75.70} & \underline{75.86} & \textbf{80.77} & \textbf{80.84} & 77.17 & 77.31 \\
  \bottomrule
\end{tabular}}
\vspace{-0.2cm}
\end{table*}

Considering that no prior work has looked into domain adaptive out-of-context news detection, we select SOTA baseline models from a similar task: domain adaptive fake news detection. 
For fair comparison, we use the same multimodal feature encoder for all baselines.

\textbf{Source only} serves as the naive baseline for the chosen backbone feature encoder. It uses one linear layer and the softmax function as the classifier. During training, it only uses the source domain.

\textbf{Source and target} uses the same architecture as \textbf{source only}. The difference is that, during training, it uses all domains, including both the source and target domains. Therefore, it can be viewed as the upper bound performance.

\textbf{EANN}~\cite{wang2018eann} is a multimodal fake news detection model that uses adversarial learning to derive event-invariant features which benefit fake news detection on newly arrived events. 

\textbf{MDA-WS}~\cite{li2021multi} uses adversarial learning to learn the domain-invariant feature, and incorporates prior knowledge to assign pseudo labels to target domain news for weak supervision.

\textbf{CANMD}~\cite{yue2022contrastive}
firstly pre-trains the model on the source training data and then generates pseudo labels for the target training data to adapt the model. 

\textbf{REAL-FND}~\cite{mosallanezhad2022domain}
uses news representation as the state and the classifiers' losses as the reward, and trains an RL agent that learns a domain-invariant news representation. 

\textbf{CADA}~\cite{li2023improving} proposes class-based adversarial domain adaptive framework to achieve fine-grained alignment.


\section{Result and Analysis}
\label{sec:result-and-analysis}

Based on the above experimental setup, we evaluate our proposed ConDA-TTA and investigate the contribution of each model component afterwards. We also include TSNE visualization and parameter sensitivity analysis (Appendix~\ref{sec:sensitivity-analysis}).

\subsection{Experimental Result}
\label{sec:experimental-result}

Table~\ref{tab:eval} shows the evaluation performance of ConDA-TTA and the baseline models. We use F1 (\%) and accuracy (Acc\%) as the evaluation metrics as in~\citet{luo-etal-2021-newsclippings} and~\citet{biamby-etal-2022-twitter}.


\textbf{Naive Baseline and Upper Bound:} We first present our observations on the naive baseline (source only) and the supervised training (source and target) results. In our experiments, we find that using BLIP-2's multimodal representation with a plain linear classifier gives competitive performances. However, we observe around 2-10\% performance gap between the naive baseline and the supervised training. 
This suggests the need for developing domain adaptation techniques to mitigate the gap.

The domain adaptation gaps of different settings vary. For news topics in Twitter-COMMs, adapting to Military Vehicles (M) has the largest domain adaptation gap (10.6\% in F1), whilst adapting to Climate Change (Cl) and Covid-19 (Cv) have relatively smaller domain adaptation gaps. We conjecture that the differences are potentially caused by the different news styles. We manually check a batch of 10 randomly sampled news from each topic. We find that Cl and Cv tend to include images that have more abundant information, containing such as embedded texts, numbers, and even news screenshots. However, in M, the news image usually contains one clear subject: the vehicle itself. At the same time, it requires more specialized knowledge to determine whether the image pairs the text within this topic. 

For news agencies in NewsCLIPpings, adapting to BBC (B) and USA Today (U) have a larger domain adaptation gap (6.81\% and 4.43\% in accuracy), while adapting to the Guardian (G) and the Washington Post (W) have a relatively smaller gap. Although the gap is not consistently large, it could cause a big difference on news platforms where a large amount of news is posted everyday. 
Thus, we believe that it is crucial to leverage MLLM feature encoders and further mitigating its domain adaptation gap with a small training cost. 

\textbf{Baseline Comparison:} We next compare our model's performance with the baseline models. On both datasets, our proposed ConDA-TTA outperforms the baseline models across most metrics. Notably, on Twitter-COMMs, ConDA-TTA outperforms the best baseline model by 2.93\% in F1 when the source domains are Covid-19 (Cv) and Climate Change (Cl), and the target domain is Military Vehicles (M). 

On NewsCLIPpings, when the target domain is BBC (B) and USA Today (U), our model outperforms the baselines. When the target domain is Guardian (G), our model achieves the second best performance.
However, we observe that when the target domain is the Washington Post (W), all models see slightly negative transfer.
This is in line with the observations on the underlying dataset~\cite{liu2020visual}, where the authors find that training their image captioning model on either the B or G results in a comparatively low CIDEr score~\citep{vedantam2015cider} on the W. As such, we conjecture that the information contained in B and G is not enough for enhancing the model's domain adaptation performance on W.

\begin{table*}
  \caption{Evaluation results on ablating $\mathcal{L}_{ctr}$, MMD and TTA of ConDA-TTA.}
  \label{tab:ablation}
  \resizebox{\textwidth}{!}{\begin{tabular}{lp{0.5cm}p{0.5cm}p{0.5cm}p{0.5cm}p{0.5cm}p{0.5cm}p{0.5cm}p{0.5cm}p{0.5cm}p{0.5cm}p{0.5cm}p{0.5cm}p{0.5cm}p{0.5cm}}
    \toprule
    \multirow{3}{*}{Model} & \multicolumn{6}{c}{Twitter-COMMs} & \multicolumn{8}{c}{NewsCLIPpings}\\
    ~ & \multicolumn{2}{c}{Cv, Cl $\rightarrow$ M} & \multicolumn{2}{c}{Cv, M $\rightarrow$ Cl} & \multicolumn{2}{c}{Cl, M $\rightarrow$ Cv} & \multicolumn{2}{c}{U, W $\rightarrow$ B} & \multicolumn{2}{c}{U, W $\rightarrow$ G} & \multicolumn{2}{c}{B, G $\rightarrow$ U} & \multicolumn{2}{c}{B, G $\rightarrow$ W} \\
    \cmidrule(lr){2-3} \cmidrule(lr){4-5} \cmidrule(lr){6-7} \cmidrule(lr){8-9} \cmidrule(lr){10-11} \cmidrule(lr){12-13} \cmidrule(lr){14-15}
    ~ & F1 & Acc & F1 & Acc & F1 & Acc & F1 & Acc & F1 & Acc & F1 & Acc & F1 & Acc \\
    \midrule
    w/o $\mathcal{L}_{ctr}$ & 78.07 & 78.12  & 80.59 & 80.59  & \underline{80.10} & \underline{80.11} & 70.80 & 70.97 & 75.07 & 75.28 & 80.27 & 80.33 & \underline{77.33} & \underline{77.50} \\
    w/o MMD & 75.85 & 75.92 & 80.47 & 80.48 & 79.14 & 79.17 & \underline{71.38} & \underline{71.46} & \underline{75.64} & \underline{75.80} & 79.93 & 79.98 & 77.22 & 77.35\\
    w/o TTA & \underline{78.58} & \underline{78.59} & \underline{80.73} & \underline{80.73} & \textbf{80.52} & \textbf{80.58} & 70.20 & 70.26 & 75.68 & 75.68 & \underline{80.75} & \underline{80.77} & \textbf{77.89} & \textbf{77.89}\\
    \midrule
    ConDA-TTA & \textbf{79.26} & \textbf{79.34} & \textbf{80.80} & \textbf{80.80} & 80.06 & 80.08 & \textbf{71.52} & \textbf{71.80} & \textbf{75.70} & \textbf{75.86} & \textbf{80.77} & \textbf{80.84} & 77.17 & 77.31 \\
  \bottomrule
\end{tabular}}
\vspace{-0.3cm}
\end{table*}

In summary, ConDA-TTA outperforms the baselines in most of the domain adaptation settings. The improvements are more significant when the domain adaptation gap is larger.
(\{Cv, Cl $\rightarrow$ M\}, \{U, W $\rightarrow$ G\}).

\subsection{Ablation Study}
\label{sec:ablation-study}

We next present an ablation study on different components in ConDA-TTA. Specifically, we ablate $\mathcal{L}_{ctr}$, MMD and TTA from ConDA-TTA. The evaluation results are summarized in Table~\ref{tab:ablation}.

On Twitter-COMMs, we observe that ablating MMD causes the biggest performance drop in all three domain adaptation settings.
Additionally, ablating $\mathcal{L}_{ctr}$ and TTA both result in performance decrease when the target domain is Military Vehicles (M) and Climate Change (Cl).
However, we see a marginal performance increase 
when the target domain is Covid-19 (Cv). 

On NewsCLIPpings, we observe similar degradation when ablating these three components for target domain being BBC (B), Guardian (G) and USA Today (U). Specifically, when ablating TTA, the performance decreases 
most significantly on B, and less significantly on G
and U.
However, when adapting to Washington Post (W), ablating $\mathcal{L}_{ctr}$, MMD and TTA all bring slight performance increase. 
We assume that this observation echos the previous analysis (Section~\ref{sec:experimental-result}) that the information contained in BBC and Guardian is not enough for helping the domain adaptation to Washington Post, therefore making each component less effective.

In summary, the above results confirm the efficacy of $\mathcal{L}_{ctr}$, MMD and TTA under most domain adaptation scenarios on both datasets. 
Concretely, MMD tends to contribute the most on Twitter-COMMs; TTA tends to contribute the most on NewsCLIPpings. 
It is worth noting that news in Twitter-COMMs are more colloquial and informal, leading writing styles vary much across different topics. We conclude that MMD is better at capturing domain-invariant features in this scenario.

\subsection{Visualization}
\label{sec:tsne-analysis}

\begin{figure}[htbp]
  \centering
  \includegraphics[width=0.23\textwidth]{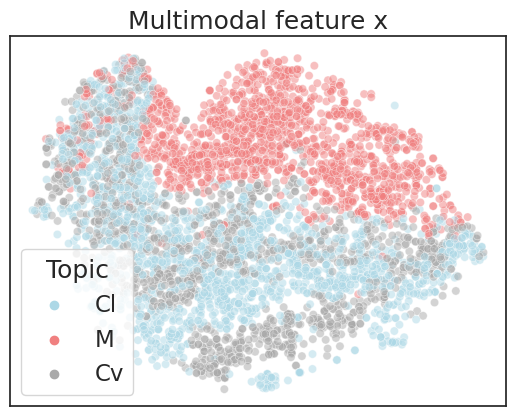}
  \includegraphics[width=0.23\textwidth]{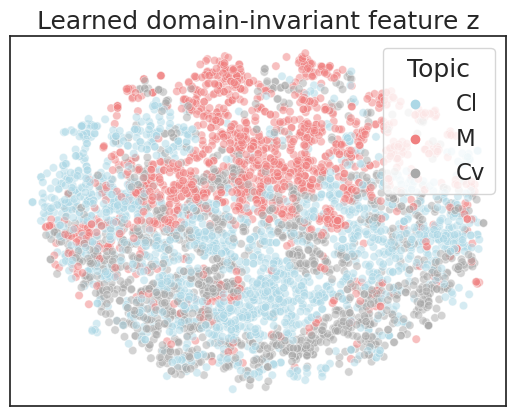}
  \caption{TSNE visualization of the multimodal feature $\textbf{x}$ and the learned domain-invariant feature $\textbf{z}$ under Cv, Cl $\rightarrow$ M.}
  \label{fig:tsne}
  \vspace{-0.2cm}
\end{figure}

To show that ConDA-TTA effectively learns the domain-invariant feature, we adopt TSNE~\cite{van2008visualizing} to visualize the multimodal feature $\textbf{x}$ 
and the learned domain-invariant feature $\textbf{z}$ projected to the 2-d space (Figure~\ref{fig:tsne}).


We can tell from the figure that feature $\textbf{z}$ of different topics blend better than $\textbf{x}$. Additionally, in Table~\ref{tab:variance}, we compute the variance of $\textbf{x}$ and $\textbf{z}$ after projecting and normalizing them onto 1-d space. The decrease of variance and the TSNE visualization
suggest that our model has removed domain-specific features from $\textbf{x}$ to some extent. This further shows that our approach effectively learns domain-invariant features.
The parameter sensitivity analysis can be found in Appendix~\ref{sec:sensitivity-analysis}.

\begin{table}
  \centering
  \caption{Variance of $\textbf{x}$ and $\textbf{z}$ on different domains.}
  \label{tab:variance}
  \resizebox{0.7\columnwidth}{!}{\begin{tabular}{lrrr}
    \toprule
    Feature & Var(M) & Var(Cl) & Var(Cv) \\
    \midrule
    $\textbf{x}$ & 0.0623 & 0.0738 & 0.0780 \\
    $\textbf{z}$ & 0.0326 & 0.0652 & 0.0724 \\
  \bottomrule
\end{tabular}}
\vspace{-0.3cm}
\end{table}

\section{Conclusion}
\label{sec:conclusion}

This paper has proposed ConDA-TTA, a domain adaptive out-of-context news detection model that can adapt to both unlabeled news topics and news agencies. ConDA-TTA first encodes the news using the BLIP-2 multimodal feature encoder and then adopts contrastive learning and MMD to learn domain-invariant features. During test-time, it further incorporates target domain statistics into the classifier.
Experimental results show the effectiveness and superiority of our approach. 

\section*{Limitations}

Despite the effectiveness of our proposed ConDA-TTA model, there are several limitations. First, in this work, we focus on two domain types: news topics and news agencies. We do not consider other potential types of domain, such as news from different regions, languages, or cultural backgrounds. Second, we observe that under some domain adaptation settings, all experimented models exhibit negative transfer. In this paper, we present analysis based on the findings in the underlying dataset paper. We will leave a deeper analysis into our future work.

\bibliography{custom}

\begin{thebibliography}{39}
\providecommand{\natexlab}[1]{#1}

\bibitem[{pyt(2024)}]{pytorch-bn}
 2024.
\newblock Batchnorm2d.
\newblock \url{https://pytorch.org/docs/stable/generated/torch.nn.BatchNorm2d.html/}.

\bibitem[{Abdelnabi et~al.(2022)Abdelnabi, Hasan, and Fritz}]{abdelnabi2022open}
Sahar Abdelnabi, Rakibul Hasan, and Mario Fritz. 2022.
\newblock Open-domain, content-based, multi-modal fact-checking of out-of-context images via online resources.
\newblock In \emph{Proceedings of the IEEE/CVF Conference on Computer Vision and Pattern Recognition}, pages 14940--14949.

\bibitem[{Bhattacharjee et~al.(2023)Bhattacharjee, Kumarage, Moraffah, and Liu}]{bhattacharjee-etal-2023-conda}
Amrita Bhattacharjee, Tharindu Kumarage, Raha Moraffah, and Huan Liu. 2023.
\newblock {C}on{DA}: Contrastive domain adaptation for {AI}-generated text detection.
\newblock In \emph{Proceedings of the 13th International Joint Conference on Natural Language Processing and the 3rd Conference of the Asia-Pacific Chapter of the Association for Computational Linguistics (Volume 1: Long Papers)}, pages 598--610, Nusa Dua, Bali. Association for Computational Linguistics.

\bibitem[{Biamby et~al.(2022)Biamby, Luo, Darrell, and Rohrbach}]{biamby-etal-2022-twitter}
Giscard Biamby, Grace Luo, Trevor Darrell, and Anna Rohrbach. 2022.
\newblock {T}witter-{COMM}s: Detecting climate, {COVID}, and military multimodal misinformation.
\newblock In \emph{Proceedings of the 2022 Conference of the North American Chapter of the Association for Computational Linguistics: Human Language Technologies}, pages 1530--1549, Seattle, United States. Association for Computational Linguistics.

\bibitem[{Chi et~al.(2023)Chi, Gu, Zhong, Liu, YU, Plataniotis, and Wang}]{chi2023adapting}
Zhixiang Chi, Li~Gu, Tao Zhong, Huan Liu, YUANHAO YU, Konstantinos~N Plataniotis, and Yang Wang. 2023.
\newblock Adapting to distribution shift by visual domain prompt generation.
\newblock In \emph{The Twelfth International Conference on Learning Representations}.

\bibitem[{Dai et~al.(2024)Dai, Li, Li, Tiong, Zhao, Wang, Li, Fung, and Hoi}]{dai2024instructblip}
Wenliang Dai, Junnan Li, Dongxu Li, Anthony Meng~Huat Tiong, Junqi Zhao, Weisheng Wang, Boyang Li, Pascale~N Fung, and Steven Hoi. 2024.
\newblock Instructblip: Towards general-purpose vision-language models with instruction tuning.
\newblock \emph{Advances in Neural Information Processing Systems}, 36.

\bibitem[{Dolhansky et~al.(2020)Dolhansky, Bitton, Pflaum, Lu, Howes, Wang, and Ferrer}]{dolhansky2020deepfake}
Brian Dolhansky, Joanna Bitton, Ben Pflaum, Jikuo Lu, Russ Howes, Menglin Wang, and Cristian~Canton Ferrer. 2020.
\newblock The deepfake detection challenge (dfdc) dataset.
\newblock \emph{arXiv preprint arXiv:2006.07397}.

\bibitem[{Fazio(2020)}]{fazio2020ooc}
Lisa Fazio. 2020.
\newblock Out-of-context photos are a powerful low-tech form of misinformation.
\newblock \url{https://theconversation.com/out-of-context-photos-are-a-powerful-low-tech-form-of-misinformation-129959/}.

\bibitem[{Gretton et~al.(2012)Gretton, Borgwardt, Rasch, Sch{\"o}lkopf, and Smola}]{gretton2012kernel}
Arthur Gretton, Karsten~M Borgwardt, Malte~J Rasch, Bernhard Sch{\"o}lkopf, and Alexander Smola. 2012.
\newblock A kernel two-sample test.
\newblock \emph{The Journal of Machine Learning Research}, 13(1):723--773.

\bibitem[{Li et~al.(2023{\natexlab{a}})Li, Wang, He, Zhang, and Liu}]{li2023improving}
Jingqiu Li, Lanjun Wang, Jianlin He, Yongdong Zhang, and Anan Liu. 2023{\natexlab{a}}.
\newblock Improving rumor detection by class-based adversarial domain adaptation.
\newblock In \emph{Proceedings of the 31st ACM International Conference on Multimedia}, pages 6634--6642.

\bibitem[{Li et~al.(2023{\natexlab{b}})Li, Chen, He, Yu, Liu, and Jia}]{li2023rethinking}
Jingyao Li, Pengguang Chen, Zexin He, Shaozuo Yu, Shu Liu, and Jiaya Jia. 2023{\natexlab{b}}.
\newblock Rethinking out-of-distribution (ood) detection: Masked image modeling is all you need.
\newblock In \emph{Proceedings of the IEEE/CVF Conference on Computer Vision and Pattern Recognition}, pages 11578--11589.

\bibitem[{Li et~al.(2023{\natexlab{c}})Li, Li, Savarese, and Hoi}]{li2023blip}
Junnan Li, Dongxu Li, Silvio Savarese, and Steven Hoi. 2023{\natexlab{c}}.
\newblock Blip-2: Bootstrapping language-image pre-training with frozen image encoders and large language models.
\newblock \emph{arXiv preprint arXiv:2301.12597}.

\bibitem[{Li et~al.(2021{\natexlab{a}})Li, Selvaraju, Gotmare, Joty, Xiong, and Hoi}]{li2021align}
Junnan Li, Ramprasaath Selvaraju, Akhilesh Gotmare, Shafiq Joty, Caiming Xiong, and Steven Chu~Hong Hoi. 2021{\natexlab{a}}.
\newblock Align before fuse: Vision and language representation learning with momentum distillation.
\newblock \emph{Advances in neural information processing systems}, 34:9694--9705.

\bibitem[{Li et~al.(2019)Li, Yatskar, Yin, Hsieh, and Chang}]{li2019visualbert}
Liunian~Harold Li, Mark Yatskar, Da~Yin, Cho-Jui Hsieh, and Kai-Wei Chang. 2019.
\newblock Visualbert: A simple and performant baseline for vision and language.
\newblock \emph{arXiv preprint arXiv:1908.03557}.

\bibitem[{Li et~al.(2016)Li, Wang, Shi, Liu, and Hou}]{li2016revisiting}
Yanghao Li, Naiyan Wang, Jianping Shi, Jiaying Liu, and Xiaodi Hou. 2016.
\newblock Revisiting batch normalization for practical domain adaptation.
\newblock \emph{arXiv preprint arXiv:1603.04779}.

\bibitem[{Li et~al.(2021{\natexlab{b}})Li, Lee, Kordzadeh, Faber, Fiddes, Chen, and Shu}]{li2021multi}
Yichuan Li, Kyumin Lee, Nima Kordzadeh, Brenton Faber, Cameron Fiddes, Elaine Chen, and Kai Shu. 2021{\natexlab{b}}.
\newblock Multi-source domain adaptation with weak supervision for early fake news detection.
\newblock In \emph{2021 IEEE International Conference on Big Data (Big Data)}, pages 668--676. IEEE.

\bibitem[{Lin et~al.(2023)Lin, Yi, Ma, Jiang, Luo, Shi, and Liu}]{lin2023zero}
Hongzhan Lin, Pengyao Yi, Jing Ma, Haiyun Jiang, Ziyang Luo, Shuming Shi, and Ruifang Liu. 2023.
\newblock Zero-shot rumor detection with propagation structure via prompt learning.
\newblock In \emph{Proceedings of the AAAI Conference on Artificial Intelligence}, volume~37, pages 5213--5221.

\bibitem[{Liu et~al.(2020)Liu, Wang, Wang, and Ordonez}]{liu2020visual}
Fuxiao Liu, Yinghan Wang, Tianlu Wang, and Vicente Ordonez. 2020.
\newblock Visual news: Benchmark and challenges in news image captioning.
\newblock \emph{arXiv preprint arXiv:2010.03743}.

\bibitem[{Long et~al.(2015)Long, Cao, Wang, and Jordan}]{long2015learning}
Mingsheng Long, Yue Cao, Jianmin Wang, and Michael Jordan. 2015.
\newblock Learning transferable features with deep adaptation networks.
\newblock In \emph{International conference on machine learning}, pages 97--105. PMLR.

\bibitem[{Luo et~al.(2021)Luo, Darrell, and Rohrbach}]{luo-etal-2021-newsclippings}
Grace Luo, Trevor Darrell, and Anna Rohrbach. 2021.
\newblock {N}ews{CLIP}pings: {A}utomatic {G}eneration of {O}ut-of-{C}ontext {M}ultimodal {M}edia.
\newblock In \emph{Proceedings of the 2021 Conference on Empirical Methods in Natural Language Processing}, pages 6801--6817, Online and Punta Cana, Dominican Republic. Association for Computational Linguistics.

\bibitem[{Ma(2019)}]{ma2019nlpaug}
Edward Ma. 2019.
\newblock Nlp augmentation.
\newblock https://github.com/makcedward/nlpaug.

\bibitem[{Mosallanezhad et~al.(2022)Mosallanezhad, Karami, Shu, Mancenido, and Liu}]{mosallanezhad2022domain}
Ahmadreza Mosallanezhad, Mansooreh Karami, Kai Shu, Michelle~V Mancenido, and Huan Liu. 2022.
\newblock Domain adaptive fake news detection via reinforcement learning.
\newblock In \emph{Proceedings of the ACM Web Conference 2022}, pages 3632--3640.

\bibitem[{Mu et~al.(2023)Mu, Das~Bhattacharjee, and Yuan}]{mu2023self}
Michael Mu, Sreyasee Das~Bhattacharjee, and Junsong Yuan. 2023.
\newblock Self-supervised distilled learning for multi-modal misinformation identification.
\newblock In \emph{Proceedings of the IEEE/CVF Winter Conference on Applications of Computer Vision}, pages 2819--2828.

\bibitem[{Pan et~al.(2010)Pan, Tsang, Kwok, and Yang}]{pan2010domain}
Sinno~Jialin Pan, Ivor~W Tsang, James~T Kwok, and Qiang Yang. 2010.
\newblock Domain adaptation via transfer component analysis.
\newblock \emph{IEEE transactions on neural networks}, 22(2):199--210.

\bibitem[{Qi et~al.(2024)Qi, Yan, Hsu, and Lee}]{qi2024sniffer}
Peng Qi, Zehong Yan, Wynne Hsu, and Mong~Li Lee. 2024.
\newblock Sniffer: Multimodal large language model for explainable out-of-context misinformation detection.
\newblock \emph{arXiv preprint arXiv:2403.03170}.

\bibitem[{Qian et~al.(2022)Qian, Li, Zhang, Jin, Fan, and Dai}]{qian2022contrastive}
Tao Qian, Fei Li, Meishan Zhang, Guonian Jin, Ping Fan, and Wenhua Dai. 2022.
\newblock Contrastive learning from label distribution: A case study on text classification.
\newblock \emph{Neurocomputing}, 507:208--220.

\bibitem[{Radford et~al.(2021)Radford, Kim, Hallacy, Ramesh, Goh, Agarwal, Sastry, Askell, Mishkin, Clark et~al.}]{radford2021learning}
Alec Radford, Jong~Wook Kim, Chris Hallacy, Aditya Ramesh, Gabriel Goh, Sandhini Agarwal, Girish Sastry, Amanda Askell, Pamela Mishkin, Jack Clark, et~al. 2021.
\newblock Learning transferable visual models from natural language supervision.
\newblock In \emph{International conference on machine learning}, pages 8748--8763. PMLR.

\bibitem[{Shalabi et~al.(2024)Shalabi, Felouat, Nguyen, and Echizen}]{shalabi2024leveraging}
Fatma Shalabi, Hichem Felouat, Huy~H Nguyen, and Isao Echizen. 2024.
\newblock Leveraging chat-based large vision language models for multimodal out-of-context detection.
\newblock \emph{arXiv preprint arXiv:2403.08776}.

\bibitem[{Shalabi et~al.(2023)Shalabi, Nguyen, Felouat, Chang, and Echizen}]{shalabi2023image}
Fatma Shalabi, Huy~H Nguyen, Hichem Felouat, Ching-Chun Chang, and Isao Echizen. 2023.
\newblock Image-text out-of-context detection using synthetic multimodal misinformation.
\newblock In \emph{2023 Asia Pacific Signal and Information Processing Association Annual Summit and Conference (APSIPA ASC)}, pages 605--612. IEEE.

\bibitem[{Tian et~al.(2020)Tian, Sun, Poole, Krishnan, Schmid, and Isola}]{tian2020makes}
Yonglong Tian, Chen Sun, Ben Poole, Dilip Krishnan, Cordelia Schmid, and Phillip Isola. 2020.
\newblock What makes for good views for contrastive learning?
\newblock \emph{Advances in neural information processing systems}, 33:6827--6839.

\bibitem[{Van~der Maaten and Hinton(2008)}]{van2008visualizing}
Laurens Van~der Maaten and Geoffrey Hinton. 2008.
\newblock Visualizing data using t-sne.
\newblock \emph{Journal of machine learning research}, 9(11).

\bibitem[{Vedantam et~al.(2015)Vedantam, Lawrence~Zitnick, and Parikh}]{vedantam2015cider}
Ramakrishna Vedantam, C~Lawrence~Zitnick, and Devi Parikh. 2015.
\newblock Cider: Consensus-based image description evaluation.
\newblock In \emph{Proceedings of the IEEE conference on computer vision and pattern recognition}, pages 4566--4575.

\bibitem[{Wang et~al.(2018)Wang, Ma, Jin, Yuan, Xun, Jha, Su, and Gao}]{wang2018eann}
Yaqing Wang, Fenglong Ma, Zhiwei Jin, Ye~Yuan, Guangxu Xun, Kishlay Jha, Lu~Su, and Jing Gao. 2018.
\newblock Eann: Event adversarial neural networks for multi-modal fake news detection.
\newblock In \emph{Proceedings of the 24th acm sigkdd international conference on knowledge discovery \& data mining}, pages 849--857.

\bibitem[{Yuan et~al.(2021)Yuan, Zheng, Ye, Qian, and Zhang}]{YUAN2021113633}
Hua Yuan, Jie Zheng, Qiongwei Ye, Yu~Qian, and Yan Zhang. 2021.
\newblock Improving fake news detection with domain-adversarial and graph-attention neural network.
\newblock \emph{Decision Support Systems}, 151:113633.

\bibitem[{Yuan et~al.(2023)Yuan, Guo, Qiu, Huang, and Li}]{yuan2023support}
Xin Yuan, Jie Guo, Weidong Qiu, Zheng Huang, and Shujun Li. 2023.
\newblock Support or refute: Analyzing the stance of evidence to detect out-of-context mis-and disinformation.
\newblock \emph{arXiv preprint arXiv:2311.01766}.

\bibitem[{Yue et~al.(2022)Yue, Zeng, Kou, Shang, and Wang}]{yue2022contrastive}
Zhenrui Yue, Huimin Zeng, Ziyi Kou, Lanyu Shang, and Dong Wang. 2022.
\newblock Contrastive domain adaptation for early misinformation detection: A case study on covid-19.
\newblock In \emph{Proceedings of the 31st ACM International Conference on Information \& Knowledge Management}, pages 2423--2433.

\bibitem[{Zhang et~al.(2021)Zhang, Qian, Fang, and Xu}]{9285217}
Huaiwen Zhang, Shengsheng Qian, Quan Fang, and Changsheng Xu. 2021.
\newblock \href {https://doi.org/10.1109/TMM.2020.3042055} {Multimodal disentangled domain adaption for social media event rumor detection}.
\newblock \emph{IEEE Transactions on Multimedia}, 23:4441--4454.

\bibitem[{Zhang et~al.(2020)Zhang, Wang, Chen, Zeng, Guo, Miao, and Cui}]{9206973}
Tong Zhang, Di~Wang, Huanhuan Chen, Zhiwei Zeng, Wei Guo, Chunyan Miao, and Lizhen Cui. 2020.
\newblock Bdann: Bert-based domain adaptation neural network for multi-modal fake news detection.
\newblock In \emph{2020 International Joint Conference on Neural Networks (IJCNN)}, pages 1--8.

\bibitem[{Zhang et~al.(2023)Zhang, Trinh, Cao, Cui, and Liu}]{zhang2023detecting}
Yizhou Zhang, Loc Trinh, Defu Cao, Zijun Cui, and Yan Liu. 2023.
\newblock Detecting out-of-context multimodal misinformation with interpretable neural-symbolic model.
\newblock \emph{arXiv preprint arXiv:2304.07633}.

\end{thebibliography}

\appendix

\section{Appendix}
\label{sec:appendix}

\subsection{Implementation Details}
\label{sec:appendix-implementation}

\textbf{Augmentation:}
The performance of contrastive learning is largely influenced by the transformation used to generate positive samples~\citep{tian2020makes}. 
Following the data augmentation selection in ~\citet{bhattacharjee-etal-2023-conda}, 
we experiment with a set of data augmentation techniques and select the one with the best performance on the test set as the final strategy. In view of the large size of the training set (Section~\ref{sec:datasets}), we randomly sample 1.5\% data items from the training set as a toy training set, roughly the same size as the test set.
We do this in order to expedite the selection of positive sample generation strategy. 
We experiment with a set of augmentations: \textbf{text-level:} \{\textit{synonym replacement}, \textit{random swap}, \textit{random crop}\}, \textbf{image-level:} \{\textit{random resize and crop}, \textit{random horizontal flip}, \textit{Gaussian blur}\},
and the combination of text-level and image-level augmentations. 

Below is the detailed descriptions of them:

\textbf{Text-level:}
\begin{itemize}[noitemsep,nolistsep]
    \item \textbf{Synonym Replacement:} Following~\cite{bhattacharjee-etal-2023-conda}, only words identified as nouns, adjectives, adverbs and verbs will be considered for synonym replacement. We randomly choose $n$ words of these parts-of-speech to be replaced, where $n=10\% \times$ the number of words in the sentence. We use \texttt{nltk.corpus.wordnet} to obtain the synonyms and randomly choose one for each original word.
    
    \item \textbf{Random Swap:} We randomly swap $n$ pair of words in the sentence, where $n=10\% \times$ the number of words in the sentence. This is achieved by using the \texttt{nlpaug} package~\cite{ma2019nlpaug}.
    
    \item \textbf{Random Crop:} We randomly remove $n$ consecutive words from the sentence, where $n=10\% \times$ the number of words in the sentence. This is achieved by using the \texttt{nlpaug} package~\cite{ma2019nlpaug}.
\end{itemize}

\textbf{Image-level:}
\begin{itemize}[noitemsep,nolistsep]
    \item \textbf{Random Resize and Crop:} We crop a random region of the image resize it to (224, 224).
    
    \item \textbf{Random Horizontal Flip:} We flip the image horizontally with a probability of 0.5.
    
    \item \textbf{Gaussian Blur:} We blur the image with the random Gaussian kernel. The kernel size range is from 5 to 9. The standard deviation range is from 0.1 to 5.0. All of the image-level augmentations are achieved by using \texttt{torchvision.transforms}.
\end{itemize}

Our experiments (Table~\ref{tab:augment}) show that \emph{Gaussian blur} results in the best overall performance. Thus, in our approach, we apply Gaussian blur to the image while keeping the text unaltered to generate the positive sample of the anchor item.

\begin{table}
  \caption{Evaluation on different augmentation strategies of ConDA-TTA on Twitter-COMMs. Performances are reported in accuracy. The best performance is in \textbf{bold text}.}
  \label{tab:augment}
  \resizebox{\columnwidth}{!}{\begin{tabular}{lrrrr}
    \toprule
    \multirow{2}{*}{Augmentation} & \multicolumn{3}{c}{Target Domain} & ~ \\
    \cmidrule{2-4}
    ~ & M & Cl & Cv & Avg\\
    \midrule
    Synonym Replacement & 79.7 & 80.2 & \textbf{80.3} & 80.07 \\
    Random Swap & 79.9 & 80.2 & \textbf{80.3} & 80.00 \\
    Random Crop & 79.3 & \textbf{80.5} & 80.0 & 79.93 \\
    Random Resize and Crop & \textbf{80.1} & 80.2 & 79.9 & 80.07 \\
    Random Horizontal Flip & 79.4 & 80.7 & 80.0 & 80.03 \\
    Gaussian Blur & 80.0 & 80.4 & 80.1 & \textbf{80.16} \\
    Synonym Replacement & \multirow{2}{*}{79.5} & \multirow{2}{*}{80.4} & \multirow{2}{*}{80.5} & \multirow{2}{*}{80.13} \\
    + Gaussian Blur & & & & \\
  \bottomrule
\end{tabular}}
\end{table}

\textbf{Multimodal Feature Encoder Selection:} We experiment with \one BLIP-2~\citep{li2023blip} multimodal and \two unimodal feature extractor, \three CLIP unimodal feature extractor and \four ALBEF~\citep{li2021align} multimodal feature extractor.
For \two and \three, we experiment with vector concatenation and element-wise multiplication to obtain the multimodal feature. 

We then train a one-layer linear classifier with the four features as inputs respectively and evaluate the performances (Table~\ref{tab:feature-encoder}).
Experimental results show that BLIP-2 multimodal features gives the best performance (in terms of accuracy) on both the training and test sets, as well as having the significant domain adaptation gap. 
We also evaluate the zero-shot performance of the BLIP-2 Image Text Matching head which only outputs a similarity score rather than the feature vector.
We find it underperforms (achieving around 0.5 in accuracy on both training and test set) using BLIP-2 multimodal feature directly.

\begin{table}
\centering
  \caption{Evaluation on different multimodal feature encoders of ConDA-TTA on Twitter-COMMs. Performances are reported in accuracy.}
  \label{tab:feature-encoder}
  \resizebox{0.9\columnwidth}{!}{\begin{tabular}{lrrr}
    \toprule
    \multirow{2}{*}{Multimodal Feature Encoder} & \multicolumn{3}{c}{Target Domain} \\
    \cmidrule{2-4}
    ~ & M & Cl & Cv \\ 
    \midrule
    BLIP-2\textsubscript{multi} (supervised) & 80.7 & 81.5 & 81.6 \\
    BLIP-2\textsubscript{multi} (source only) & 72.4 & 78.8 & 77.9 \\
    \rowcolor{gray!30}
    $\Delta$ \textit{domain adaptation gap} & 8.3 & 2.7 & 3.7 \\
    ALBEF (supervised) & 69.3 & 66.8 & 67.3 \\
    ALBEF (source only) & 63.7 & 64.3 & 66.0 \\
    \rowcolor{gray!30}
    $\Delta$ \textit{domain adaptation gap} & 5.6 & 2.5 & 1.3 \\
    CLIP (supervised) & 76.6 & 79.3 & 78.6 \\
    CLIP (source only) & 72.1 & 77.3 & 78.1 \\
    \rowcolor{gray!30}
    $\Delta$ \textit{domain adaptation gap} & 4.5 & 2.0 & 0.5 \\
    BLIP-2\textsubscript{uni} (supervised) & 72.5 & 75.0 & 74.4 \\
    BLIP-2\textsubscript{uni} (source only) & 58.8 & 71.3 & 72.1 \\
    \rowcolor{gray!30}
    $\Delta$ \textit{domain adaptation gap} & 13.7 & 3.7 & 2.3 \\
  \bottomrule
\end{tabular}}
\end{table}

\subsection{Computing Infrastructure}

Our experiments are conducted on one NVIDIA A40 GPU. Each epoch takes around 20 mins to finish.

\subsection{Sensitivity Analysis}
\label{sec:sensitivity-analysis}

The contrastive loss, MMD and TTA are three crucial components in our approach. This section briefly studies how the weights, $\lambda_{ctr}$ and $\lambda_{MMD}$, in the loss function and the batch size $b$ in TTA affect the performance of ConDA-TTA.

\begin{figure}[htbp]
  \centering
  \includegraphics[width=0.235\textwidth]{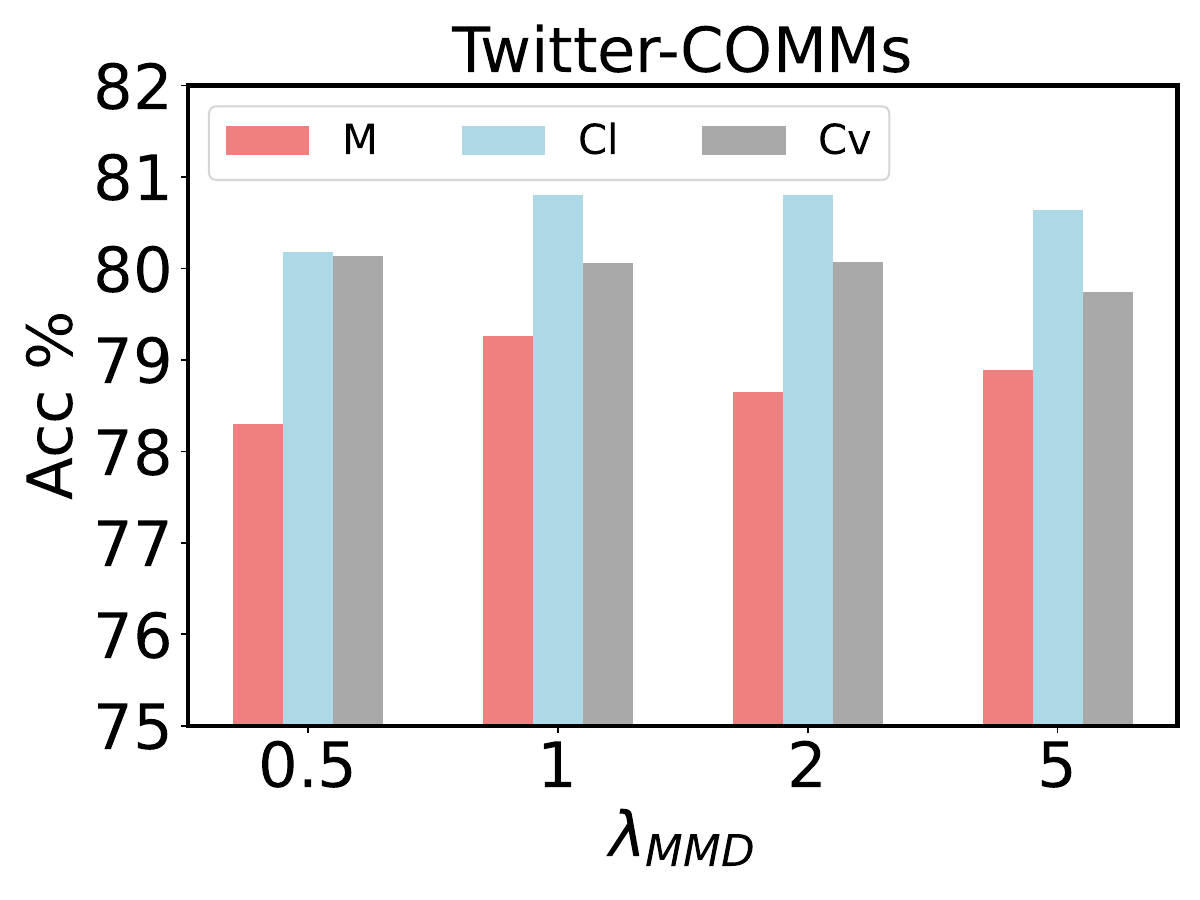}
  \includegraphics[width=0.235\textwidth]{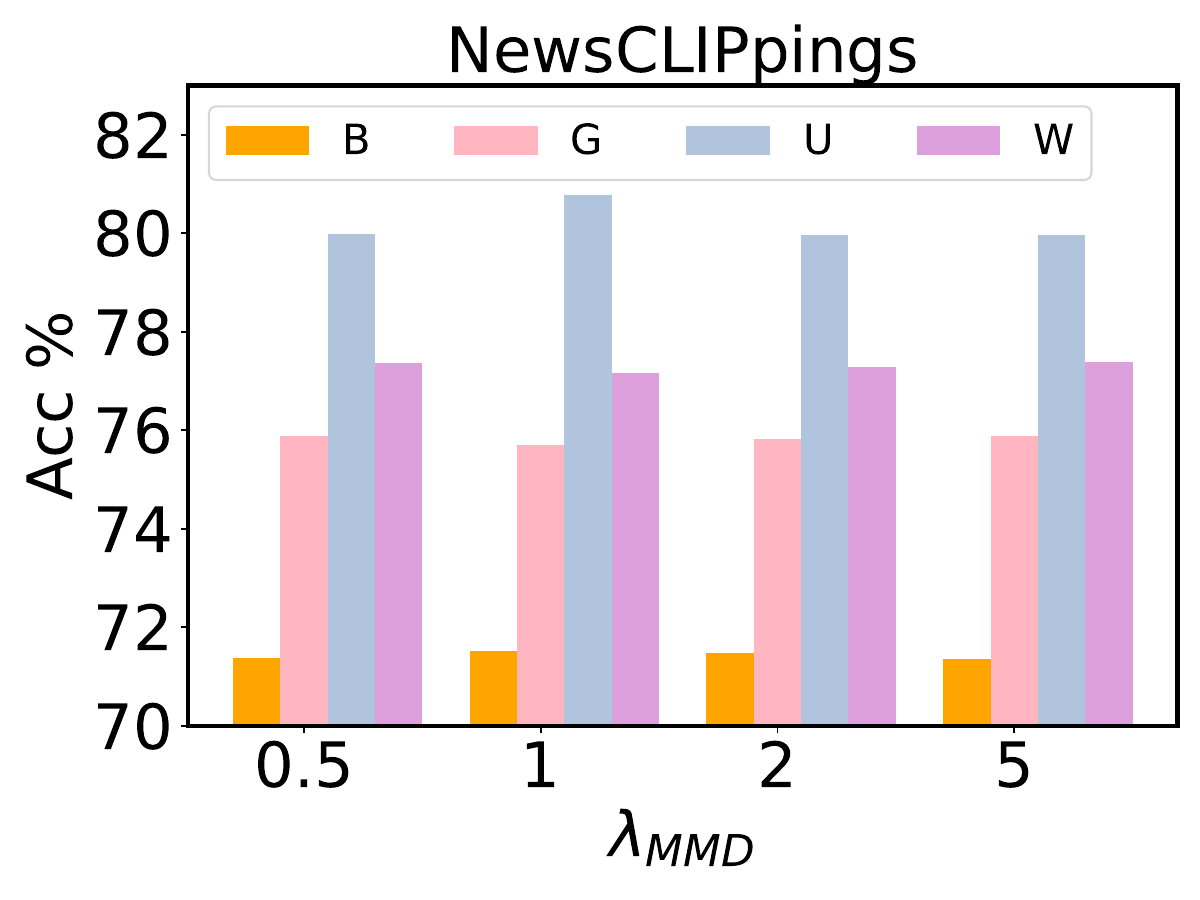}
  \includegraphics[width=0.235\textwidth]{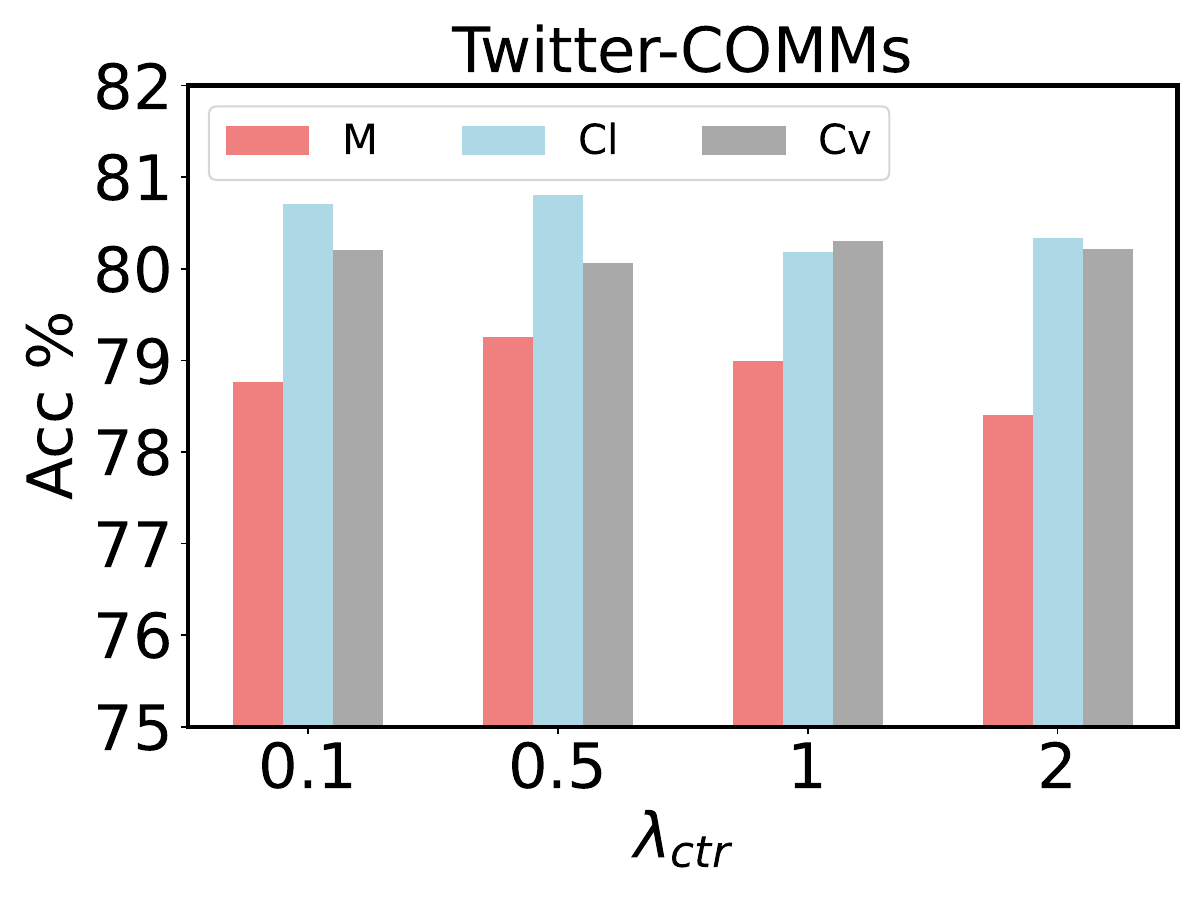}
  \includegraphics[width=0.235\textwidth]{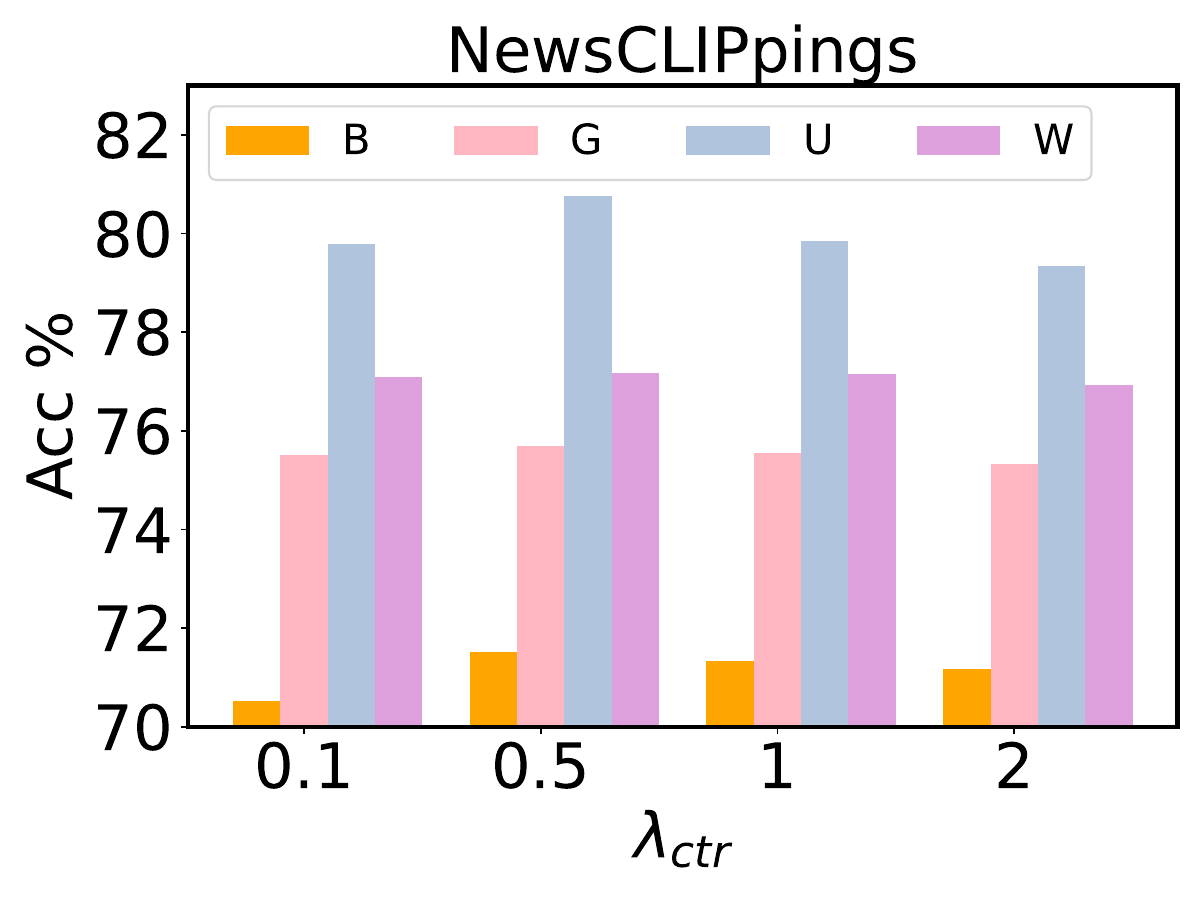}
  \caption{ConDA-TTA's performances (in Acc) with different $\lambda_{MMD}$ and $\lambda_{ctr}$ values. The legend shows the target domain.}
  \label{fig:lambda_mmd}
\end{figure}

\begin{figure}[htbp]
  \centering
  \includegraphics[width=0.235\textwidth]{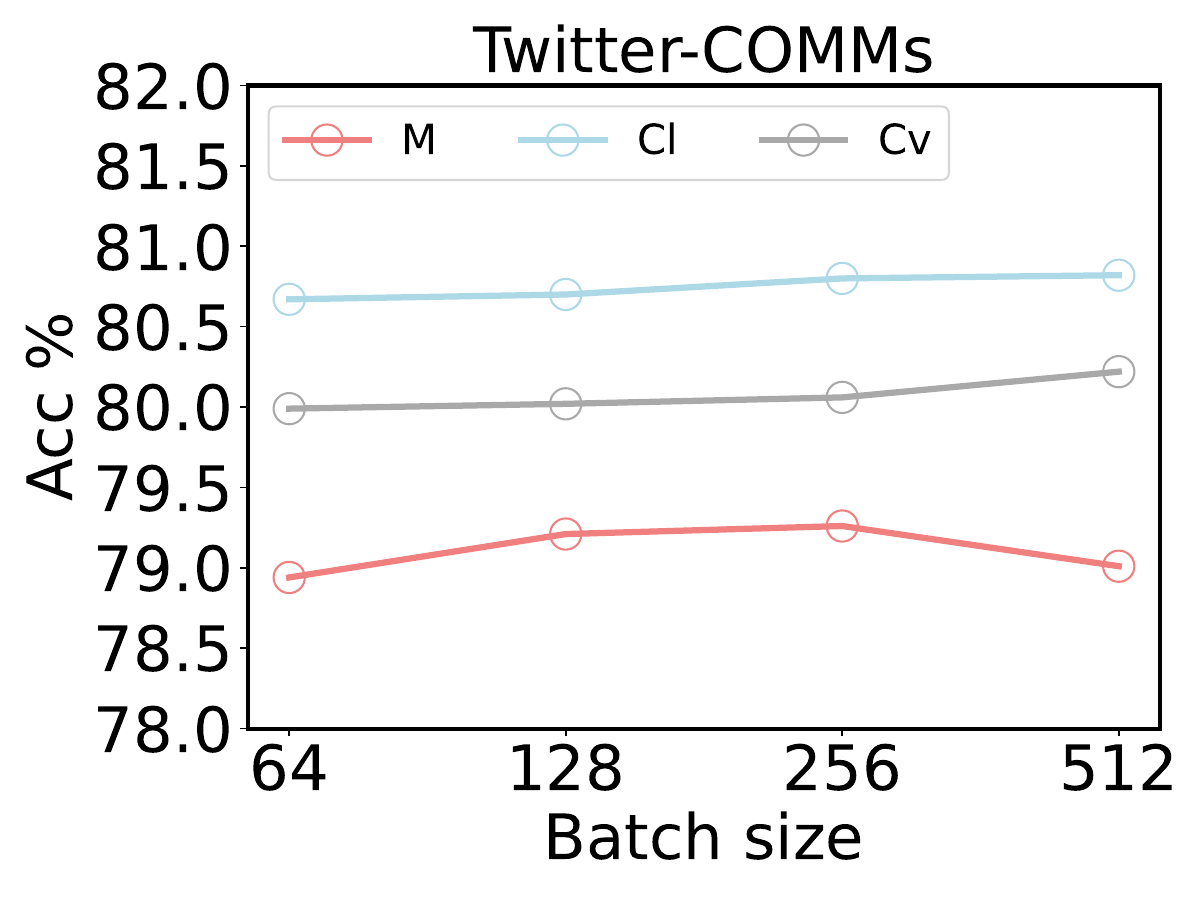}
  \includegraphics[width=0.235\textwidth]{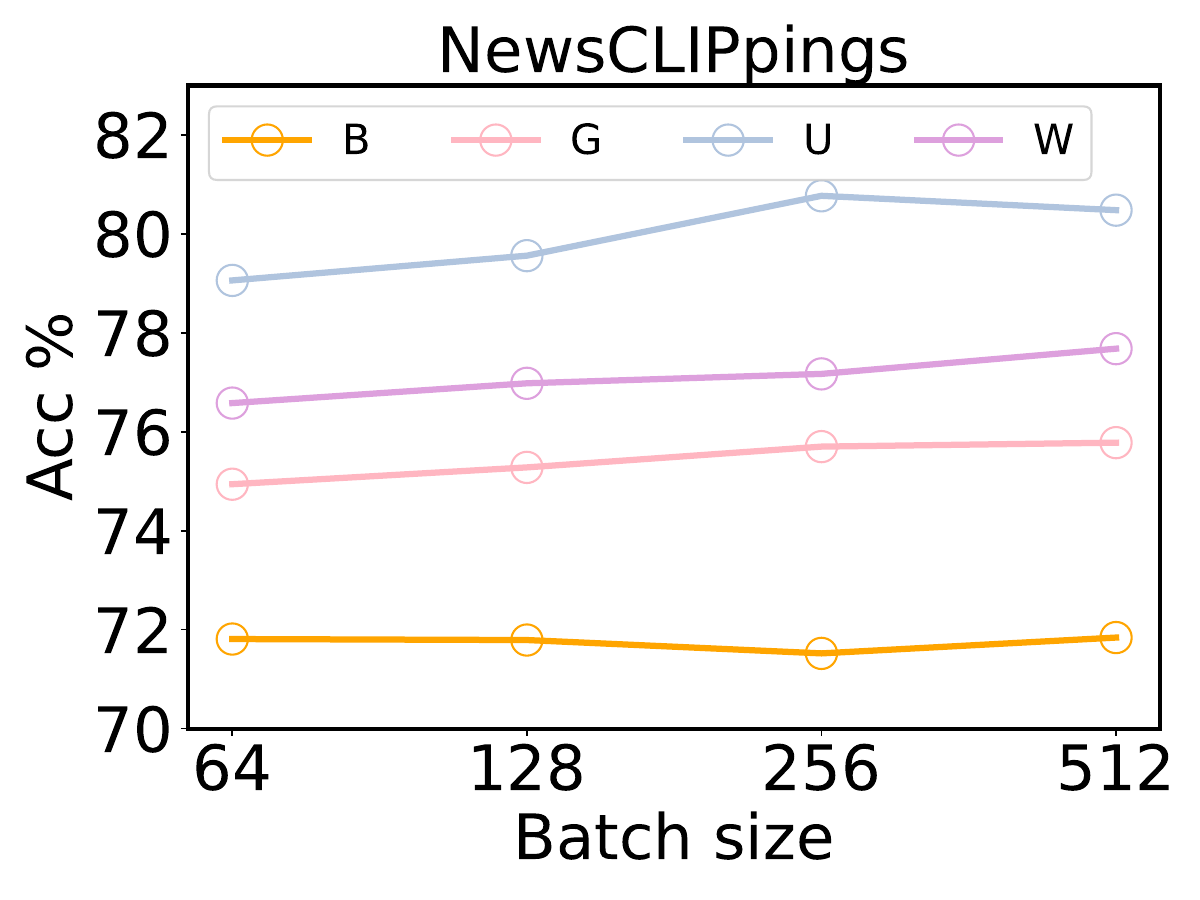}
  \caption{ConDA-TTA's performances (in Acc) with different batch sizes. The legend shows the target domain.}
  \label{fig:batch_size}
\end{figure}

Figure~\ref{fig:lambda_mmd} illustrates ConDA-TTA's performances with $\lambda_{MMD}=\{0.5, 1, 2, 5\}$ and $\lambda_{ctr}=\{0.1, 0.5, 1, 2\}$. 
When setting $\lambda_{MMD}$ to 1, our model achieves the best performance in 4 (M, Cl, B, U) out of 7 domain adaptation settings, and achieves close to best performance in the remaining 3 settings. 
Overall, $\lambda_{MMD}$ being too high causes the accuracy to drop a little. This might be because the model has put relatively less emphasis on the cross-entropy losses in Eq.~\eqref{eq:total-loss}. For $\lambda_{ctr}$, 
we find that setting it to 0.5 achieves the best performance in 5 settings. 
we find that when its value goes up, the model performance tends to decrease gradually. We conjecture this is because the distances (Section~\ref{sec:domain-invariant}) between news with the same label have become much larger, making it more difficult to classify.

Figure~\ref{fig:batch_size} illustrates ConDA-TTA's performances with test set's batch size $b=\{64, 128, 256, 512\}$. Overall, if computing resources permit, larger batch size tend to result in better model performances on most domain adaptation settings. However, the differences are not much. We argue that larger batch size helps the TTA to learn a less biased target domain statistics, because it includes more data points in one batch. However, we also conclude that it's influence is limited.
\end{document}